\documentclass{article} 
\usepackage{nips13submit_e,times}
\usepackage{hyperref}
\usepackage{url}
\usepackage{graphicx}
\usepackage{amsmath,amsfonts,amssymb}
\usepackage{bm}
\usepackage{xcolor}

\def\B#1{\boldsymbol{#1}}

\title{Mapping cognitive ontologies to and from the brain}

\author{
Yannick Schwartz, Bertrand Thirion, and Gael Varoquaux \\
Parietal Team, Inria Saclay Ile-de-France \\
Saclay, France \\
\texttt{firstname.lastname@inria.fr} \\
}

\nipsfinalcopy 

\begin{document}

\maketitle

\begin{abstract}
Imaging neuroscience links brain activation maps to behavior and
cognition via correlational studies. Due to the nature of the
individual experiments, based on eliciting neural response from a small 
number of
stimuli, this link is incomplete, and unidirectional from the causal
point of view. To come to conclusions on the function implied 
by the activation of brain regions, it is necessary to combine
a wide exploration of the various brain
functions and some inversion of the statistical inference. 
Here we introduce a methodology for accumulating knowledge
towards a bidirectional link between observed brain activity and the
corresponding function. We rely on a large corpus of imaging studies and
a predictive engine. Technically, the challenges are to find commonality
between the studies without denaturing the richness of the corpus.
The key elements that we contribute are labeling the tasks
performed with a cognitive ontology, and modeling the long tail of rare
paradigms in the corpus. To our knowledge, our approach is the first
demonstration of predicting the cognitive content of completely new brain
images. To that end, we propose a method that predicts
the experimental paradigms across different studies.
\end{abstract}

\section{Introduction}

Functional brain imaging, in particular fMRI, is the workhorse of brain
mapping, the systematic study of which areas of the brain are recruited
during various experiments. To date, 33K papers on pubmed mention
``fMRI'', revealing an accumulation of activation maps related to
specific tasks or cognitive concepts. From this literature has emerged the
notion of brain modules specialized to a task, such as the celebrated 
fusiform face area (FFA) dedicated to face recognition \cite{Kanwisher1997}.
However, the link between the brain
images and high-level notions from psychology is mostly done manually, due
to the lack of co-analysis framework. The challenges in quantifying
observations across experiments, let alone at the level of the
literature, leads to incomplete pictures and well-known fallacies. For
instance a common trap is that of \emph{reverse inferences}
\cite{poldrack2006}: attributing
a cognitive process to a brain region, while the individual experiments
can only come to the conclusion that it is recruited by the
process under study, and not that the observed activation of the region
demonstrates the engagement of the cognitive process. Functional
specificity can indeed only be measured by probing a large variety of functions,
which exceeds the scale of a single study. Beyond this lack of
specificity, individual studies are seldom comprehensive, in the sense
that they do not recruit every brain region.

Prior work on such large scale cognitive mapping of the brain has mostly
relied on coordinate-based meta-analysis, that forgo activation maps and
pool results across publications via the reported Talairach coordinates of
activation foci \cite{laird2005,yarkoni2011}. While the
underlying thresholding of statistical maps and extraction of local
maxima leads to a substantial loss of information, the value of this
approach lies in the large amount of studies covered: Brainmap
\cite{laird2005}, that relies on manual analysis of the literature,
comprises 2\,298 papers, while Neurosynth \cite{yarkoni2011}, that uses
text mining, comprises 4\,393 papers. Such large corpuses can be used to
evaluate the occurrence of the cognitive and behavioral terms associated
with activations and formulate reverse inference as a Bayesian inversion
on standard (forward) fMRI inference \cite{poldrack2006,yarkoni2011}. On
the opposite end of the spectrum, \cite{poldrack2009} shows that using a
machine-learning approach on studies with different cognitive content can
predict this content from the images, thus demonstrating principled
reverse inference across studies. Similarly, \cite{hanson2008} have used
image-based classification to challenge the vision that the FFA is by
itself specific of faces. Two trends thus appear in the quest for
explicit correspondences between brain regions and cognitive concepts.
One is grounded on counting term frequency on a large corpus of studies
described by coordinates. The other uses predictive models on images.
The first approach can better define functional specificity by
avoiding the sampling bias inherent to small groups of studies;
however each study in a coordinate-based meta-analysis brings only very
limited spatial information \cite{salimi2009}.


Our purpose here is to outline a strategy to accumulate knowledge from a
brain functional image database in order to provide grounds for
principled bidirectional reasoning from brain activation to behavior and
cognition. To increase the breadth in co-analysis
and scale up from \cite{poldrack2009}, which used only 8 studies with 22
different cognitive concepts, we have to tackle several challenges. A first
challenge is to find commonalities across studies, without which we face
the risk of learning idiosyncrasies of the protocols. For this very reason
we choose to describe studies with terms that come from a cognitive paradigm
ontology instead of a high-level cognitive process one. This setting 
enables not only to span the terms across all
the studies, but also to use atypical studies that do not clearly share
cognitive processes. A second challenge is that
of diminishing statistical power with increasing number of cognitive
terms under study. Finally, a central goal is to ensure some sort of
functional specificity, which is hindered by the data scarcity and
ensuing biases in an image database.

In this paper, we gather 19 studies, comprising 131 different
conditions, which we labeled with 19 different terms describing
experimental paradigms. We perform a brain mapping experiment across
these studies, in which we consider both forward and reverse inference. Our
contributions are two-fold: on the one hand we show empirical results
that outline specific difficulties of such co-analysis, on the second
hand we introduce a methodology using image-based classification and a
cognitive-paradigm ontology that can scale to large set of studies.
The paper is organized as following. In section \ref{sec:methods}, we
introduce our methodology for establishing correspondence between studies
and performing forward and reverse inference across them. In section
\ref{sec:data}, we present our data, a corpus of studies and the
corresponding paradigm descriptions. In section \ref{sec:results} we show
empirically that our approach can predict these descriptions in unseen
studies, and that it gives promising maps for brain mapping. Finally, in
section \ref{sec:discussion}, we discuss the empirical findings in the
wider context of meta-analyses.





\section{Methodology: annotations, statistics and learning}

\label{sec:methods}

\subsection{Labeling activation maps with common terms across studies}

A standard task-based fMRI study results in \emph{activation maps} per
subject that capture the brain response to each experimental condition.
They are combined to single out responses to high-level cognitive
functions in so-called \emph{contrast maps}, for which the inference is
most often performed at the group level, across subjects. These contrasts
can oppose different experimental conditions, some to capture the effect
of interest while others serve to cancel out non-specific effects. For
example, to highlight computation processes, one might contrast
\emph{visual calculation} with \emph{visual sentences}, to suppress the
effect of the stimulus modality (visual instructions), and the explicit
stimulus (reading the numbers). 

When considering a corpus of different studies, finding correspondences
between the effects highlighted by the contrasts can be challenging.
Indeed, beyond classical \emph{localizers}, capturing only very wide
cognitive domains, each study tends to investigate fairly unique
questions, such as syntactic structure in language rather than language
in general \cite{pallier2011}. Combining the studies requires engineering
\emph{meta-contrasts} across studies. For this purpose, we choose to
affect a set of \emph{terms} describing the content of each condition.
Indeed, there are important ongoing efforts in cognitive science and
neuroscience to organize the scientific concepts into formal ontologies
\cite{turner2012cognitive}. Taking the ground-level objects of these
gives a suitable family of terms, a \emph{taxonomy} to describe the
experiments.



\subsection{Forward inference: which regions are recruited by tasks
containing a given term?}

Armed with the term labels, we can use the standard fMRI analysis
framework and ask using a General Linear Model (GLM) across studies for
each voxels of the subject-level activation images if it is
significantly-related to a term in the corpus of images. 
If $\B{x} \in \mathbb{R}^p$ is the observed activation map with $p$
voxels, the GLM tests $\mathcal{P}(\B{x}_i \neq 0 | T)$ for each voxel
$i$ and term $T$. 
This test relies on a linear model that assumes that the response in
each voxel is a combination of the different factors and on classical
statistics:
\[
\B{x} = \B{Y}\B{\beta} + \varepsilon,
\]
where $\B{Y}$ is the design matrix yielding the occurrence of terms
and $\B{\beta}$ the term effects.
Here, we assemble \emph{term-versus-rest} contrasts, that test for the
specific effect of the term.
The benefit of the GLM formulation is that it estimates the effect of
each term partialing out the effects of the other terms, and thus
imposes some form of functional specificity in the results. Term
co-occurrence in the corpus can however lead to collinearity of the
regressors.


\subsection{Reverse inference: which regions are predictive of tasks
containing a given term?}

Poldrack 2006 \cite{poldrack2006} formulates reverse inferences as
reasoning on $\mathcal{P}(T|\B{x})$, the probability of a term $T$
being involved in the experiment given the activation map $\B{x}$. For
coordinate-based meta analysis, as all that is available is the
presence or the absence of significant activations at a given position,
the information on $\B{x}$ boils down to $\{i, \B{x}_i \neq
0\}$. Approaches to build a reverse inference framework upon this
description have relied on Bayesian inversion to go from
$\mathcal{P}(\B{x}_i \neq 0 | T)$, as output by the GLM, to
$\mathcal{P}(T |\B{x}_i \neq 0)$ \cite{poldrack2006,yarkoni2011}.
In terms of predictive models on images, this approach can be
understood as a naive Bayes predictor: the distribution of the different
voxels are learned independently conditional to each term, and Bayes' rule 
is used for prediction. Learning voxels-level parameters independently is
a limitation as it makes it harder to capture distributed effects, such
as large-scale functional networks, that can be better predictors of
stimuli class than localized regions \cite{hanson2008}. However, learning
the full distribution of $\B{x}$ is ill-posed, as $\B{x}$ is
high-dimensional. For this reason, we must resort to statistical learning
tools.

We choose to use an $\ell_2$-regularized logistic regression to directly estimate the
conditional probability $\mathcal{P}(T|\B{x})$ under a linear model. The
choice of linear models is crucial to our brain-mapping goals, as their
decision frontier is fully represented by a brain map\footnote{In this regard, the Naive Bayes prediction
strategy does yield clear cut maps, as its decision boundary is a conic
section.}  $\B{\beta} \in \mathbb{R}^p$. However, as the images are spatially smooth,
neighboring voxels carry similar information, and we use feature
clustering with spatially-constrained Ward clustering
\cite{michel2012supervisedclustering} to reduce the dimensionality of the
problem. We further reduce the dimensionality by selecting the most
significant features with a one-way ANOVA. We observe that the 
classification performance is not hindered if we reduce the data 
from 48K voxels to 15K parcels\footnote{Reducing even further down to 2K
parcels does not impact the classification performance, however the brain
maps $\B{\beta}$ are then less spatially resolved.}
and then select the 30\% most significant features.
The classifier is quite robust to these parameters, and our choice
is motivated by computational concerns. We indeed use a leave-one-study
out cross validation scheme, nested with a 10-fold stratified shuffle 
split to set the $\ell_2$ regularization parameter. As a result, we need
to estimate 1200 models per term label, which amounts to over 20K in total.
The dimension reduction helps making the approach
computationally tractable.

The learning task is rendered difficult by the fact that it is highly
multi-class, with a small number of samples in some classes. To divide
the problem in simpler learning tasks, we use the fact that our terms are
derived from an ontology, and thus can be grouped by parent category. In
each category, we apply a strategy similar to one-versus-all: we train a
classifier to predict the presence of each term, opposed to the others.
The benefits of this approach are \emph{i)} that it is suited to the
presence of multiple terms for a map, and \emph{ii)} that the features it
highlights are indeed selective for the associated term only.

Finally, an additional challenge faced by the predictive learning task is
that of strongly imbalanced classes: some terms are very frequent, while
others hardly present. In such a situation, an empirical risk minimizer
will mostly model the majority class. Thus we add sample weights inverse
of the population imbalance in the training set. This strategy is
commonly used to compensate for covariate shift \cite{shimodaira2000}.
However, as our test set is drawn from the same corpus, and thus shows
the same imbalance, we apply an inverse bias in the decision rule of the
classifier by shifting the probability output by the logistic model: if
$P$ is the probability of the term presence predicted by the logistic, we
use: $P_\text{biased} = \rho_\text{term} P$, where $\rho_\text{term}$ is
the fraction of train samples containing the term.


\section{An image database}

\label{sec:data}

\subsection{Studies}

We need a large collection of task fMRI datasets to cover the cognitive space.
We also want to avoid particular biases regarding imaging methods or scanners, 
and therefore prefer images from different teams. We use 19 studies, mainly
drawn from the OpenfMRI project \cite{poldrack2013openfmri}, which despite 
remaining small in comparison to coordinate databases, is as of now the largest
open database for task fMRI. The datasets include risk-taking
tasks \cite{schonberg2012decreasing, tom2007neural}, classification 
tasks \cite{aron2006long, foerde2006modulation, poldrack2001interactive},
language tasks \cite{xue2007neural, pallier2011, vagharchakian2012},
stop-signal tasks \cite{xue2008common},
cueing tasks \cite{kelly2008competition},
object recognition tasks \cite{haxby2001, duncan2009consistency},
functional localizers tasks \cite{pinel2007, pinel2013},
and finally a saccades \& arithmetic task \cite{knops2009}. 
The database accounts for 486 subjects, 131 activation map types, and 3\,826
individual maps, the number of subjects and map types varying across the
studies. To avoid biases due to heterogeneous data analysis procedures,
we re-process from scratch all the studies with the SPM (Statistical Parametric
Mapping) software.

\subsection{Annotating}

To tackle highly-multiclass problems,
computer vision greatly benefits from the WordNet ontology \cite{deng2010} to
standardize annotation of pictures, but also to impose structure on the
classes. The neuroscience
community recognizes the value of such vocabularies and develops
ontologies to cover the different aspects of the field such as protocols, 
paradigms, brain regions and cognitive processes. Among the many initiatives,
CogPO (The Cognitive Paradigm Ontology) \cite{turner2012cognitive} aims to
represent the cognitive paradigms used in fMRI studies. CogPO focuses on the
description of the experimental conditions characteristics, namely the 
explicit stimuli and their modality, the instructions, and the explicit
responses and their modality. Each of those categories use standard terms
to specify the experimental condition. As an example a stimulus modality may
be \emph{auditory} or \emph{visual}, the explicit stimulus a 
\emph{non-vocal sound} or a \emph{shape}. We use this ontology to label with
the appropriate terms all the experimental conditions from the database. The 
categories and terms that we use are listed in Table~\ref{table:cogpo}.

\begin{table}[t]
\begin{tabular}{ll}
\multicolumn{1}{c}{\bf CATEGORY}  &\multicolumn{1}{c}{\bf TERMS}
\\ \hline \\[-1em]
Stimulus modality         &visual, auditory \\
Explicit stimulus         &words, shapes, digits, abstract patterns,
                           non-vocal sounds, scramble, face\\
Instructions             &attend, read, move, track, count, discriminate, inhibit\\
Overt response           &saccades, none, button press\\
\end{tabular}
\caption{Subset of CogPO terms and categories that are present in our
corpus}
\label{table:cogpo}
\end{table}

\section{Experimental results}

\label{sec:results}

\subsection{Forward inference}

In our corpus, the occurrence of some terms is too correlated and gives
rise to co-linear regressors. For instance, we only have visual or
auditory stimulus modalities. While a handful of contrasts display both
stimulus modalities, the fact that a stimulus is not auditory mostly amounts
to it being visual. For this reason, we exclude from our forward
inference \emph{visual}, which will be captured by negative effects on
\emph{auditory}, and \emph{digits}, that amounts mainly to the
instruction being \emph{count}.
We fit the GLM using a design matrix comprising all the remaining terms, 
and consider results with p-values corrected
for multiple comparisons at a 5\% family-wise error rate (FWER).
To evaluate the spatial layout of the different CogPO categories, we
report the different term effects as outlines in the brain, and show
the 5\% top values for each term to avoid clutter in 
Figure~\ref{fig:inference_atlas}. Forward inference 
outlines many regions relevant to the terms, such as the primary visual and
auditory systems on the \emph{stimulus modality} maps, or pattern and 
object-recognition areas in the ventral stream, on the 
\emph{explicit stimulus} maps. 



It can be difficult to impose a functional specificity in forward inference
because of several phenomena: 
\textit{i)} the correlation present in the design matrix, makes
it hard to separate highly associated (often anti-correlated) factors,
as can be seen in Fig. \ref{fig:l2distance}, right. 
\textit{ii)} the assumption inherent to this model that a certain
factor is expressed identically across all experiments where it is
present. This assumption ignores modulations and interactions effects
that are very likely to occur; however their joint occurrence is
related to the protocol, making it impossible to disentangle these
factors with the database used here. \emph{iii)} important confounding
effects are not modeled, such as the effect of attention.
Indeed the \emph{count} map captures networks related to visuo-spatial orientation and 
attention: a dorsal attentional network, and a salience network (insulo-cingulate
network \cite{seeley2007}) in Figure~\ref{fig:inference_atlas}.

\subsection{Reverse inference}

\begin{figure}[t]
\begin{minipage}{.45\linewidth}
    \centerline{\includegraphics[width=.9\linewidth]{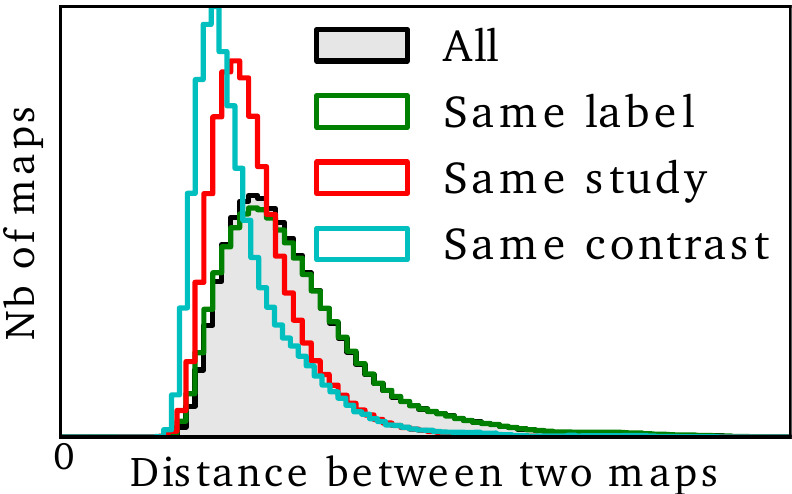}}

    \caption{(Left) Histogram of the distance between maps owing to their
    commonalities: study of origin, functional labels, functional
    contrast. (Right) Correlation of the design matrix.}
\label{fig:l2distance}
\end{minipage}%
\begin{minipage}{.56\linewidth}
    \includegraphics[width=\linewidth]{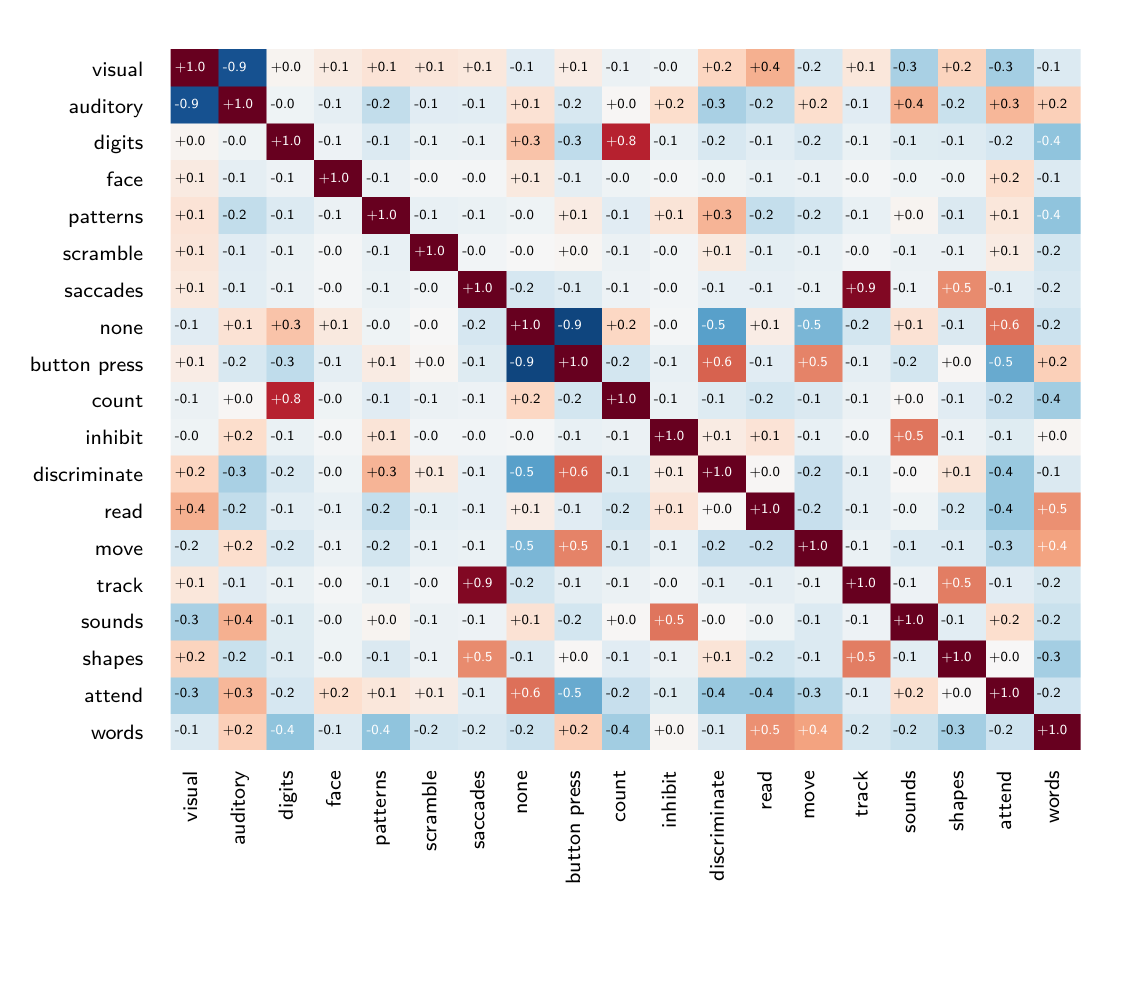}%
\end{minipage}%
\end{figure}

The promise of predictive modeling on a large statistical map database is
to provide principled reverse inference, going from observations of
neural activity to well-defined cognitive processes.
The classification model however requires a careful setting to be specific
to the intended effect. Figure~\ref{fig:l2distance} highlights 
some confounding effects that can captured by a predictive model:
two statistical
maps originating from the same study are closer than two maps labeled
as sharing a same experimental condition in the sense of a Euclidean distance.
We mitigate the capture of undesired effect with different strategies.
First we use term labels at span across studies, and refrain from using 
those that
were not present in at least two. We ensure this way that no term is
attached to a specific study. Second, we only test the classifiers on
previously unseen studies and if possible subjects, using for example
a leave-one-study out cross validation scheme. A careless
classification setting can very easily lead to training a study detector.

\begin{figure}[t]
\begin{center}
    \includegraphics[width=.85\linewidth]{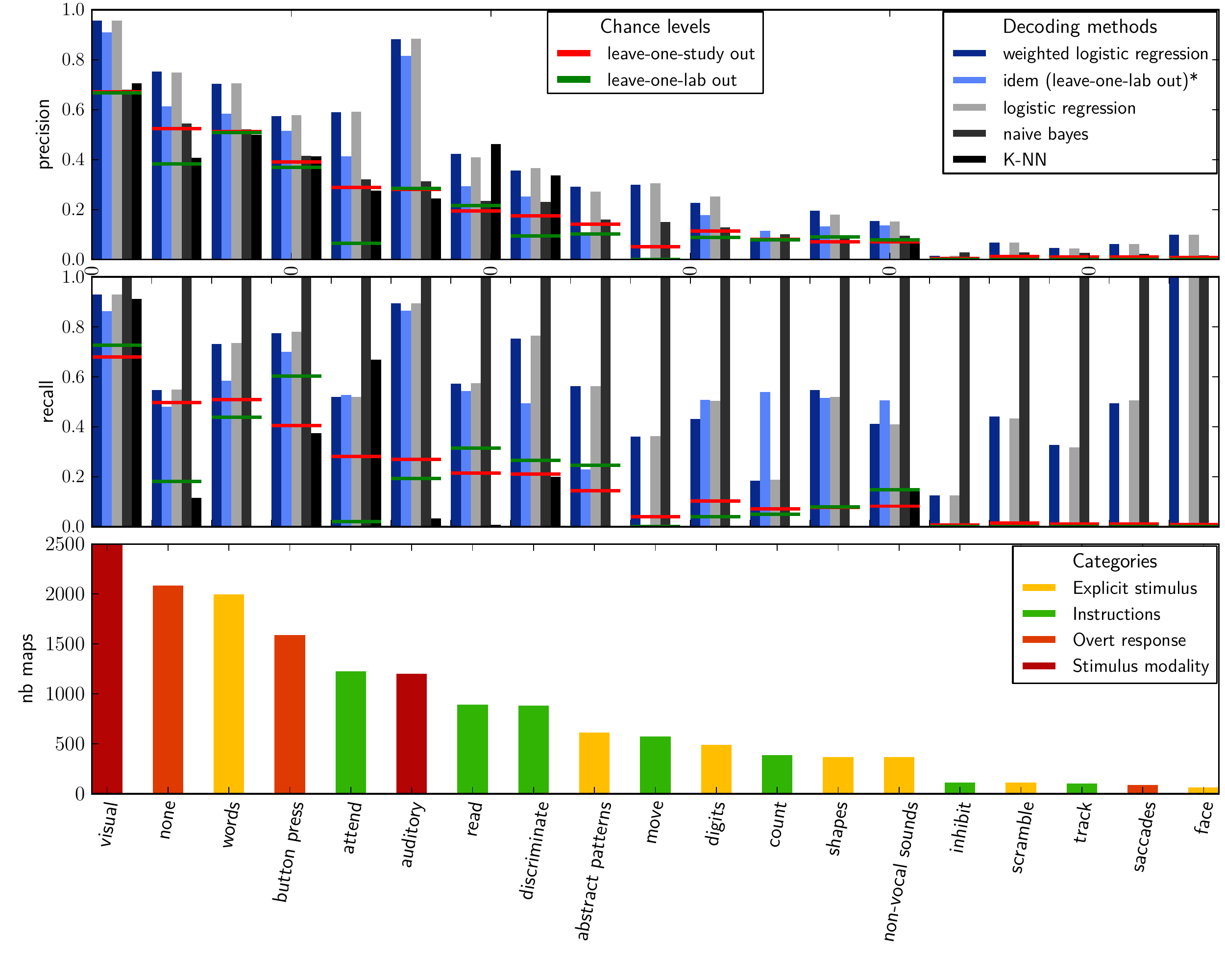}%
\end{center}
\caption{Precision and recall for all terms per classification method, 
and term representation in the database. The * denotes a leave-one-laboratory out
cross validation scheme, associated with the green bars representing the chance levels.
The other methods use a leave-one-study out cross validation, whose chance levels are
represented by the red horizontal bars.}
\label{fig:methods}
\end{figure}

Figure~\ref{fig:methods} summarizes the highly multi-class and imbalanced
problem that we face: the distribution of the number of samples per class
displays a long tail. To find non-trivial effects we need to be
able to detect the under-represented terms as well as possible. As a 
reference method, we use a K-NN, as it is in general a fairly good
approach for highly multi-class problems. Its training is independent of
the term label structure and predicts the map labels instead. It 
subsequently assigns to a new map terms that are present in more than
half of its nearest neighbors from the training\footnote{K was chosen in
a cross-validation loop, varying between 5 and 20. Such small numbers for
K are useful to avoid penalizing under-represented terms of rare classes 
in the vote of the KNN. For this reason we do not explore above K=20, 
in respect to the small number of occurrences of the \emph{faces} term.}. 
We compare this approach to training
independent predictive models for each term and use three types of classifiers:
a naive Bayes, a logistic regression, and a weighted logistic regression.
Figure~\ref{fig:methods} shows the results for each method in terms of
precision and recall, standard information-retrieval metrics.
Note that the performance scores
mainly follow the class representation, \emph{i.e.} the number of samples per
class in the train set.
Considering that rare occurrences are also those that are most
likely to provide new insight, we want a model that promotes recall over
precision in the tail of the term frequency distribution. On the other
hand, well represented classes are easier to detect and correspond to
massive, well-known mental processes. For these, we want to favor
precision, \emph{i.e.} not affecting the corresponding term to 
other processes, as these term are fairly general and non-descriptive.

Overall the K-NN has the worst performance, both in
precision and recall. It confirms the idea outlined in 
Figure~\ref{fig:l2distance}, that an Euclidean distance alone is not
appropriate to discriminate underlying brain functions because of overwhelming
confounding effects\footnote{Note that the picture does not change when $\ell_1$
distances are used instead of $\ell_2$ distances.}. 
Similarly, the naive bayes performs poorly, with very high recall and 
low precisions scores which lead to a lack of functional specificity.
On the contrary, the methods using a logistic regression show better results,
and yield performance scores above the chance levels which are represented
by the red horizontal bars for the leave-one-study out cross validation scheme in Figure~\ref{fig:methods}.
Interestingly, switching the cross validation scheme to a leave-one-laboratory out
does not change the performance significantly. This result is important, as
it confirms that the classifiers do not rely on specificities from the
stimuli presentation in a research group to perform the prediction.
We mainly use data drawn from 2 different groups in this work, and use
those data in turn to train and test a logistic regression model. The predicitions
scores for the terms present in both groups are displayed in Figure~\ref{fig:methods},
with the chance levels represented by the green horizontal bars for this cross validation
scheme.

\begin{figure}[h]
    \includegraphics[width=.54\linewidth]{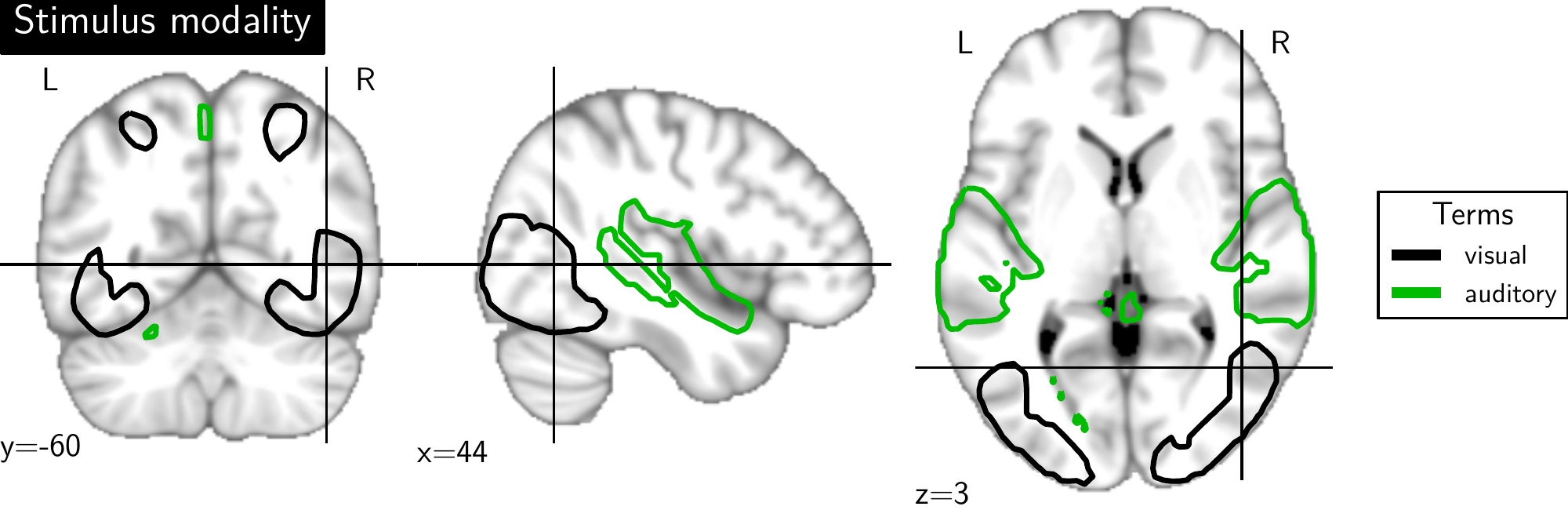}%
    \includegraphics[width=.54\linewidth]{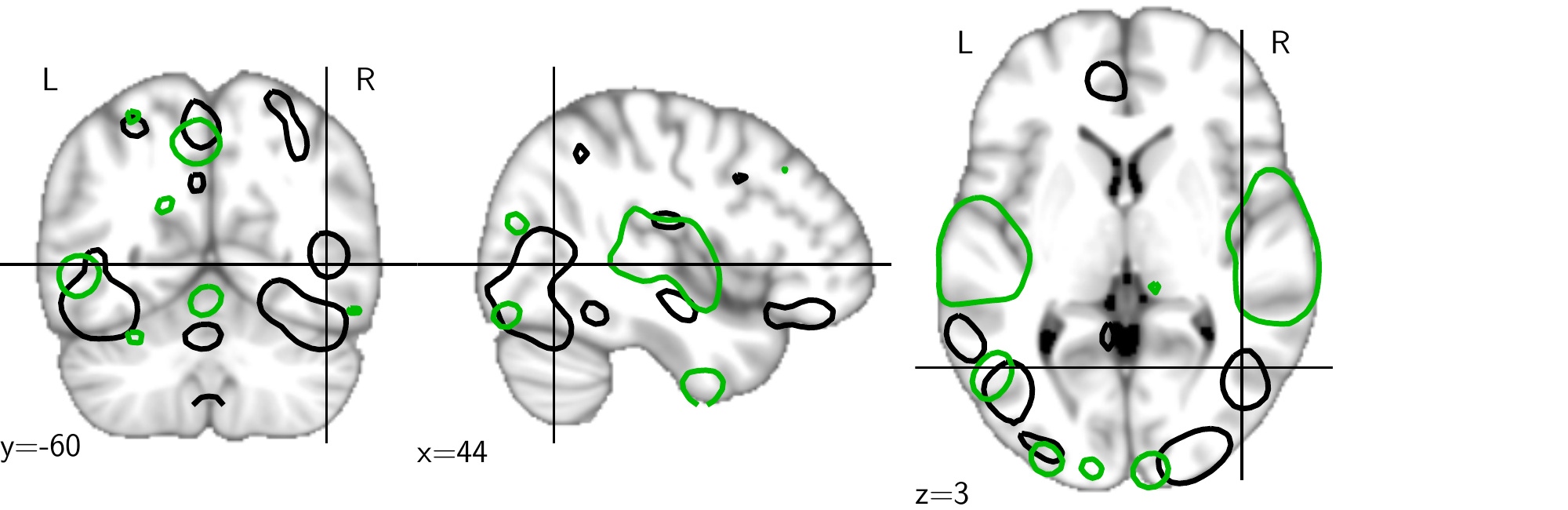}%

    \includegraphics[width=.54\linewidth]{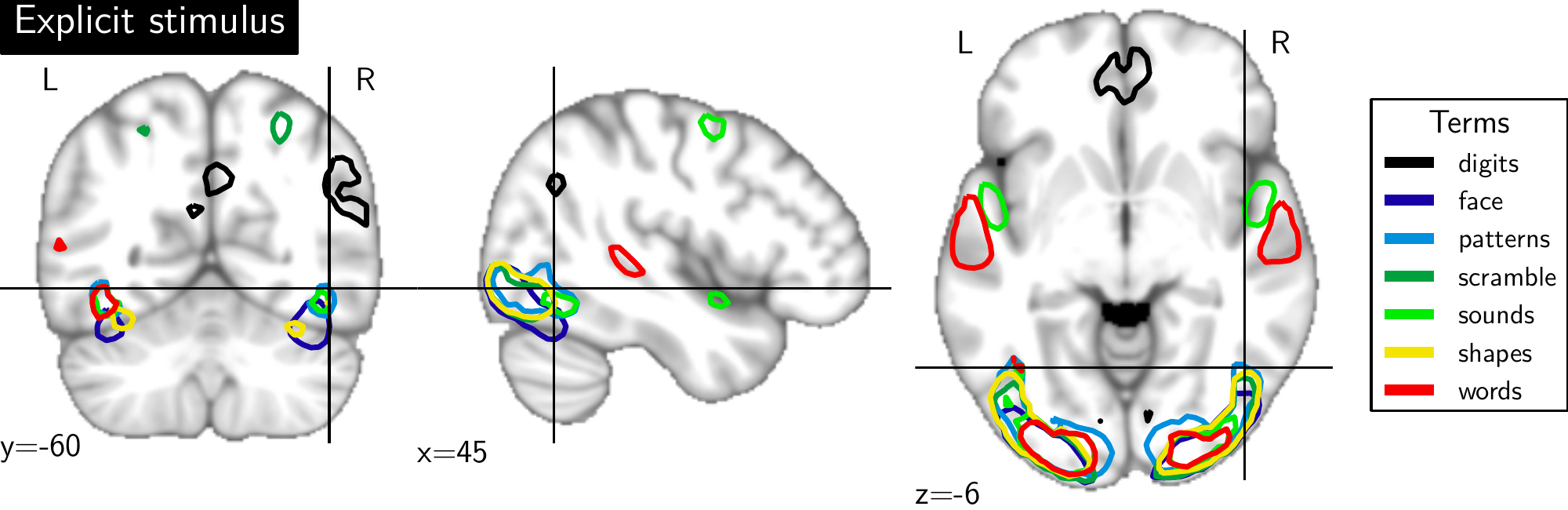}%
    \includegraphics[width=.54\linewidth]{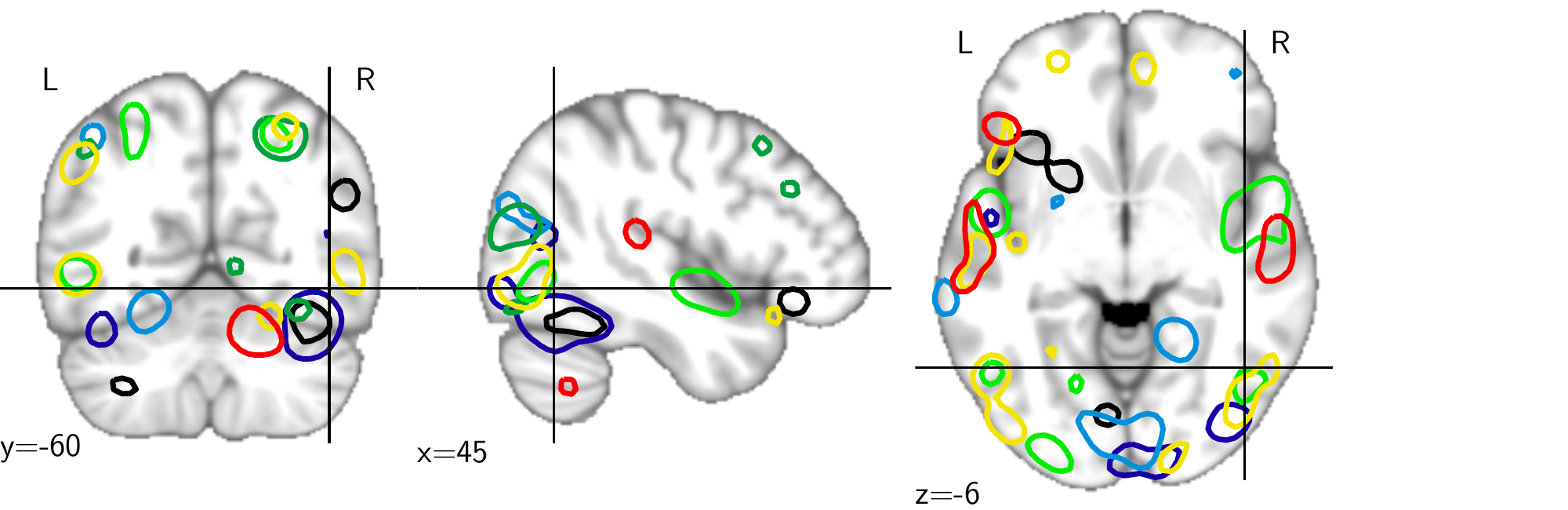}
    \vspace*{-1.3em} 

    \includegraphics[width=.54\linewidth]{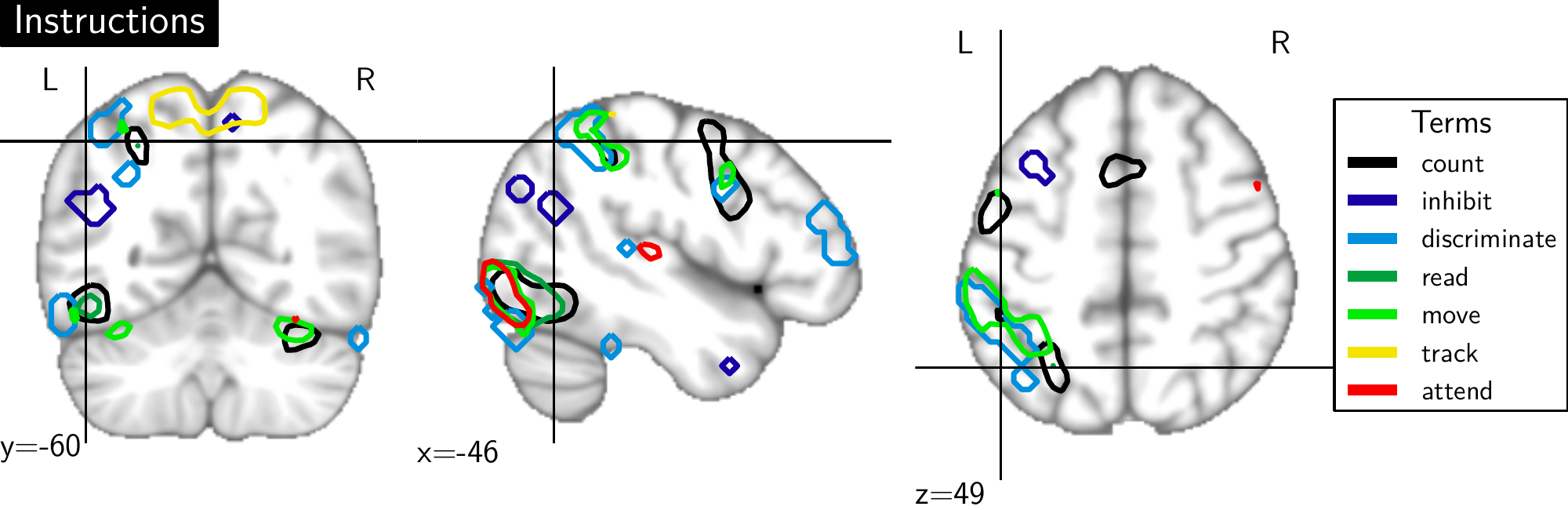}%
    \includegraphics[width=.54\linewidth]{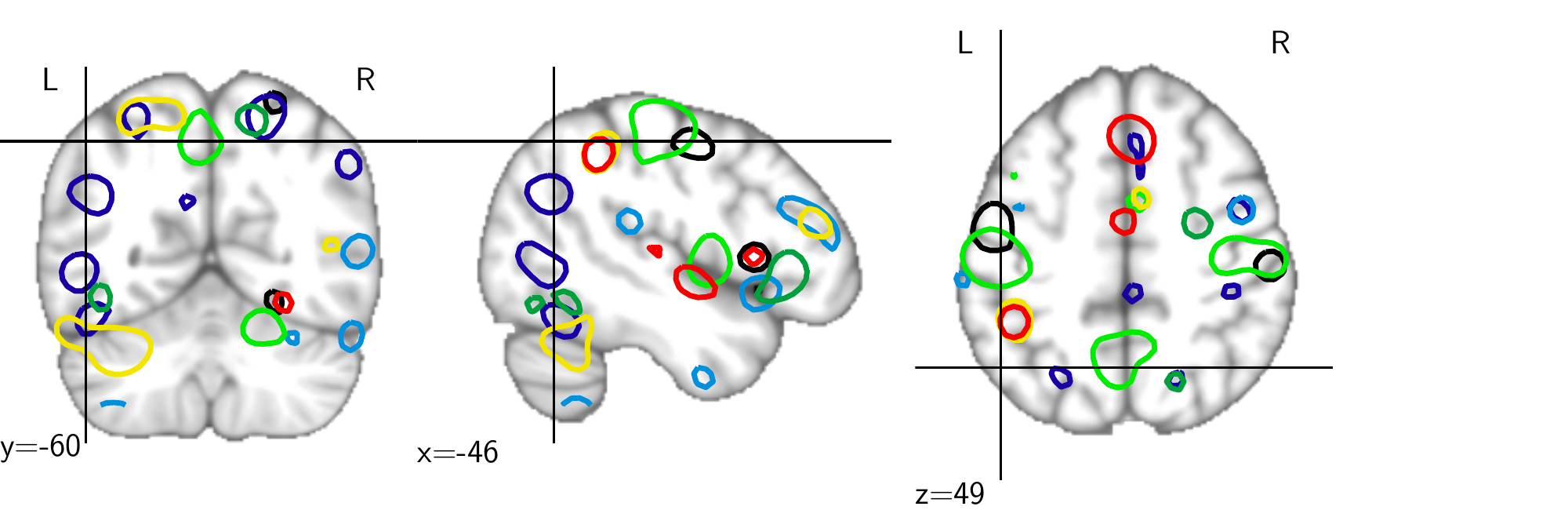}
    \vspace*{-1.3em} 
 
    \includegraphics[width=.54\linewidth]{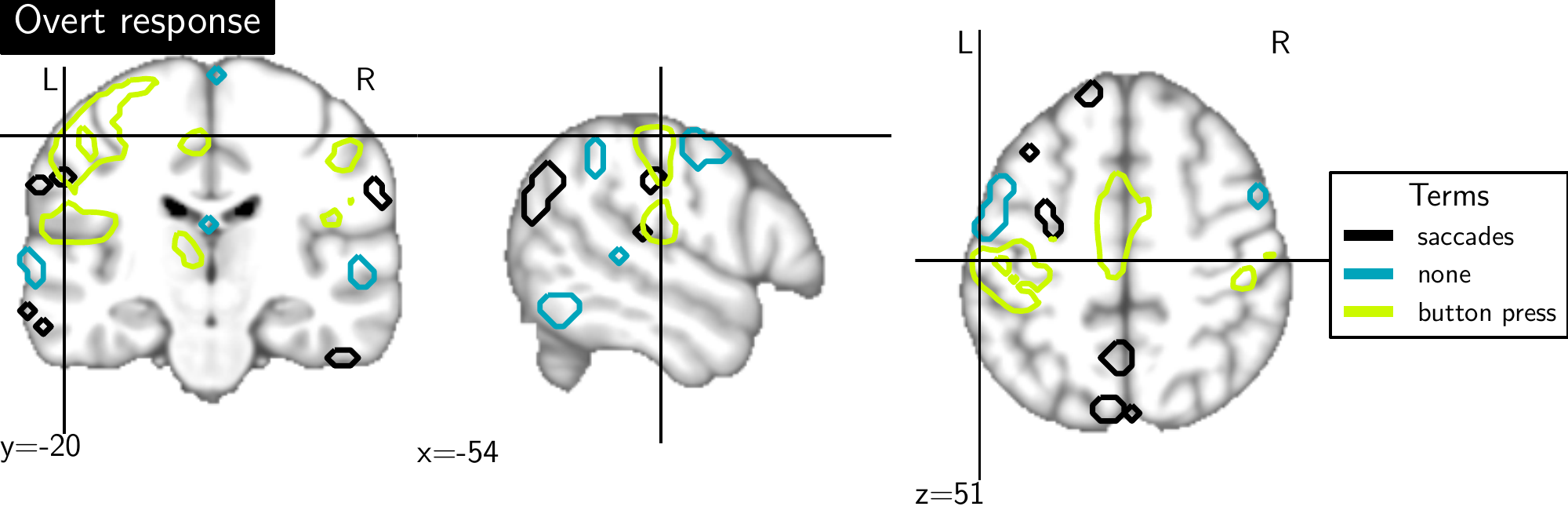}%
    \includegraphics[width=.54\linewidth]{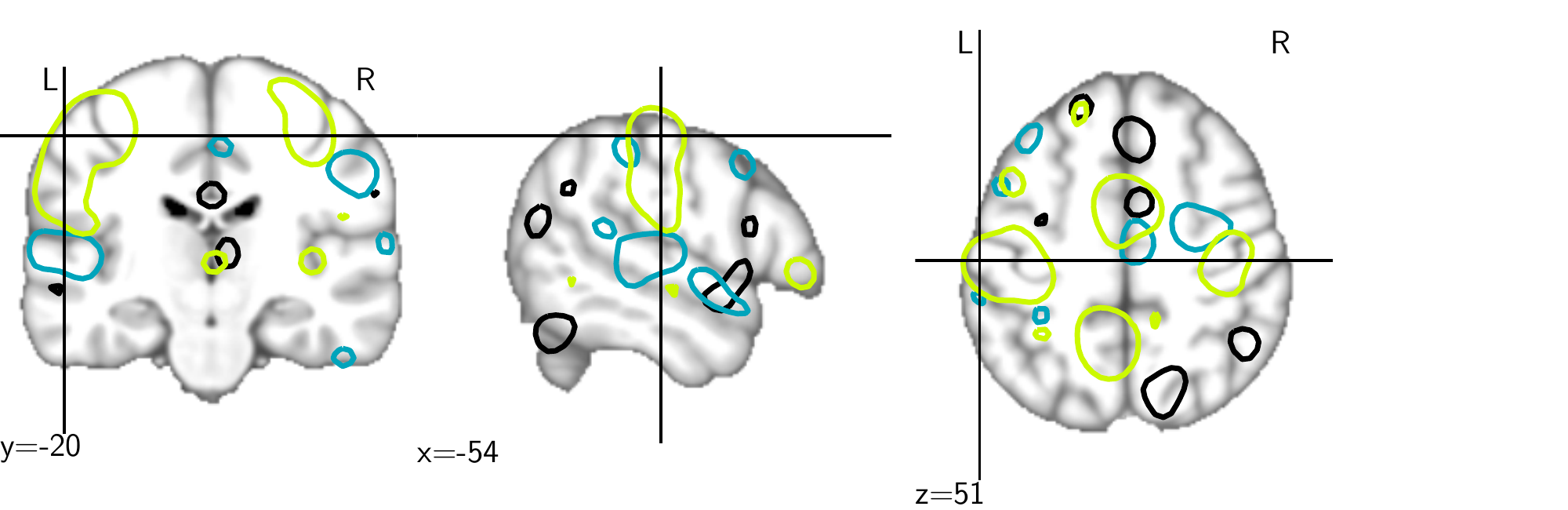}%

    \mbox{\sffamily\bfseries\rlap{~ Forward inference atlas}%
    \hspace*{.54\linewidth}~ Reverse inference atlas}
\caption{Maps for the forward inference (left) and the reverse inference
(right) for each term category. To minimize clutter, we set the outline so 
as to encompass 5\% of the voxels in the brain on each figure, thus
highlighting only the salient features of the maps. In reverse
inference, to reduce the visual effect of the parcellation, maps were smoothed
using a $\sigma$ of 2 voxels.
}
\label{fig:inference_atlas}
\end{figure}

We evaluate the spatial layout of maps representing CogPO categories for 
reverse inference as
well, and report boundaries of the 5\% top values from the weighted 
logistic coefficients. Figure~\ref{fig:inference_atlas} reports the outlined
regions that include motor cortex activations in the 
\emph{instructions} category, and activations in the auditory cortex and FFA
respectively for the words and faces terms in the \emph{explicit stimulus} category.
Despite being very noisy, those regions report findings consistent with 
the literature and complementary to the forward inference maps. For
instance, the \emph{move} instruction map comprises the motor cortex,
unlike for forward inference. Similarly, the \emph{saccades} over
response map segments the intra-parietal sulci and the frontal eye
fields, which corresponds to the well known signature of saccades, unlike
the corresponding forward inference map, which is very non specific of
saccades\footnote{This failure of forward inference is probably due to the
small sample size of saccades.}.

\section{Discussion and conclusion}

\label{sec:discussion}


Linking cognitive concepts to brain maps can give solid grounds to the
diffuse knowledge derived in imaging neuroscience. Common studies provide
evidence on which brain regions are recruited in given tasks. However
coming to conclusions on the tasks in which regions are specialized
requires data accumulation across studies to overcome the small coverage
in cognitive domain of the tasks assessed in a single study.
In practice, such a program faces a variety of roadblocks. Some are
technical challenges, that of build a statistical predictive engine that
can overcome the curse of dimensionality. While others are core to
meta-analysis. Indeed, finding correspondence between studies is a key
step to going beyond idiosyncrasies of the experimental designs. Yet the
framework should not discard rare but repeatable features of the
experiments as these provide richness to the description of brain
function.

We rely on ontologies to solve the correspondence problem. It is an
imperfect solution, as the labeling is bound to be inexact, but it brings
the benefit of several layers of descriptions and thus enable us to
fraction the multi-class learning task in simpler tasks. A similar
strategy based on WordNet was essential to progress in object recognition
in the field of computer vision \cite{deng2010}. Previous work
\cite{poldrack2009} showed high classification scores for several mental
states across multiple studies, using cross-validation with a
leave-one-subject out strategy. However, as this work did not model
common factors across studies, the mental state was confounded by the
study. In every study, a subject was represented by a single statistical
map, and there is therefore no way to validate whether the study or the
mental state was actually predicted. As figure \ref{fig:l2distance}
shows, predicting studies is much easier albeit of little neuroscientific
interest. Interestingly, \cite{poldrack2009} also explores the ability of
a model to be predictive on two different studies sharing the same
cognitive task, and a few subjects. When using the common subjects, their
model performs worse than without these subjects, as it partially
mistakes cognitive tasks for subjects. This performance drop
illustrates that a classifier
is not necessarily specific to the desired effect, and in this case
detects subjects in place of tasks to a certain degree. To avoid this
loophole, we included in our corpus only studies that had terms in
common with at least on other study and performed cross-validation by
leaving a study out, and thus predicting from completely new activation
maps. The drawback is that it limits directly the number of terms that
we can attempt to predict given a database, and explain why we have fewer
terms than \cite{poldrack2009} although we have more than twice as 
many studies. Indeed, in \cite{poldrack2009}, the terms cannot be
disambiguated from the studies.

Our labeled corpus is riddled with very infrequent terms giving rise to
class imbalance problems in which the rare occurrences are the most
difficult to model. Interestingly, though coordinates databases such as
Neurosynth \cite{yarkoni2011} cover a larger set of studies and a broader
range of cognitive processes, they suffer from a similar imbalance bias,
which is given by the state of the literature. Indeed, by looking at the
terms in Neurosynth, that are the closest to the one we use in this work,
we find that \emph{motor} is cited in 1090 papers, \emph{auditory} 558,
\emph{word} 660, and the number goes as low as 55 and 31 for
\emph{saccade} and \emph{calculation} respectively. Consequently, these
databases may also yield inconsistent results. For instance, the reverse
inference map corresponding to the term \emph{digits} is empty, whereas the
forward inference map is well defined
\footnote{\url{http://neurosynth.org/terms/digits}}. Neurosynth draws
from almost 5K studies while our work is based on 19 studies; however,
unlike Neurosynth, we are able to benefit from the different contrasts 
and subjects in our studies, which provides us with 3\,826 training
samples. In this regard, our approach is particularly interesting and can
hope to achieve competitive results with much less studies.

This paper shows the first demonstration of \emph{zero-shot learning} for
prediction of tasks from brain activity: paradigm description is given
for images from unseen studies, acquired on different scanners, in
different institutions, on different cognitive domains. More importantly
than the prediction per se, we pose the foundation of a framework 
to integrate and co-analyze many studies. This data accumulation,
combined with the predictive model can provide good proxies of 
\emph{reverse inference maps}, giving regions whose activation supports 
certain cognitive functions. These maps should, in principle, be better suited for
causal interpretation than maps estimated from standard
brain mapping correlational analysis.
In future work, we plan to control the significance of the reverse 
inference maps, that show promising results but would probably benefit from
thresholding out non-significant regions. In addition, we hope that
further progress, in terms of spatial and cognitive resolution in
mapping the brain to cognitive ontologies, will come from enriching 
the database with new studies, that will bring more images, and new
low and high-level concepts.


\subsubsection*{Acknowledgments}

This work was supported by the ANR grants BrainPedia ANR-10-JCJC 1408-01
and IRMGroup ANR-10-BLAN-0126-02, as well as the NSF grant NSF OCI-1131441
for the OpenfMRI project.

\small
\bibliographystyle{ieeetr}

\bibliography{biblio}

\end{document}